\def\paperID{2105}
\def\confName{CVPR}
\def\confYear{2024}

\def\paperTitle{SIGNeRF: Scene Integrated Generation for Neural Radiance Fields}

\def\authorBlock{
Jan-Niklas Dihlmann\qquad
Andreas Engelhardt\qquad
Hendrik Lensch \\
University of T{\"u}bingen \\
{\tt\small \{jan-niklas.dihlmann, andreas.engelhardt, hendrik.lensch\}@uni-tuebingen.de}
}

\newif\ifreview 
\newif\ifarxiv \newcommand{\arxiv}{\arxivtrue}
\newif\ifcamera 
\newif\ifrebuttal 

\arxiv %

\pdfoutput=1
\documentclass[10pt,twocolumn,letterpaper]{article}

\ifreview \usepackage[review]{cvpr} \fi
\ifarxiv \usepackage[pagenumbers]{cvpr} \fi
\ifrebuttal \usepackage[rebuttal]{cvpr} \fi
\ifcamera \usepackage{cvpr} \fi

\usepackage{graphicx}
\usepackage{amsmath}
\usepackage{amssymb}
\usepackage{booktabs}
\usepackage{times}
\usepackage{microtype}
\usepackage{epsfig}
\usepackage[table,xcdraw,dvipsnames]{xcolor}
\usepackage{caption}
\usepackage{float}
\usepackage{placeins}
\usepackage{color, colortbl}
\usepackage{stfloats}
\usepackage{enumitem}
\usepackage{tabularx}
\usepackage{xstring}
\usepackage{multirow}
\usepackage{xspace}
\usepackage{url}
\usepackage{subcaption}
\usepackage{xcolor}
\usepackage[hang,flushmargin]{footmisc}

\usepackage{siunitx}
\usepackage{array}
\usepackage{makecell}

\ifcamera \usepackage[accsupp]{axessibility} \fi

\ifarxiv  \fi

\definecolor{r1}{HTML}{2F729D}
\definecolor{r2}{HTML}{9A6B56}
\definecolor{r3}{HTML}{3F795A}

\newcommand{\R}[1]{{%
            \textbf{%
                \ifstrequal{#1}{1}{\textcolor{r1}{R#1}}{%
                    \ifstrequal{#1}{2}{\textcolor{r2}{R#1}}{%
                        \ifstrequal{#1}{3}{\textcolor{r3}{R#1}}{%
                            \ifstrequal{#1}{4}{\textcolor{teal}{R#1}}{%
                                \textcolor{cyan}{R#1}%
                            }}}}%
            }%
        }}

\usepackage{xr-hyper}

\makeatletter
\newcommand*{\addFileDependency}[1]{
  \typeout{(#1)}
  \@addtofilelist{#1}
  \IfFileExists{#1}{}{\typeout{No file #1.}}
}

\makeatother

\definecolor{cvprblue}{rgb}{0.21,0.49,0.74}
\usepackage[pagebackref,breaklinks,colorlinks,citecolor=cvprblue]{hyperref}
\usepackage[capitalize]{cleveref}
\crefname{section}{Sec.}{Secs.}
\crefname{table}{Table}{Tables}
\crefname{figure}{Fig.}{Figs.}

\title{\paperTitle}
\author{\authorBlock}
\begin{document}

\twocolumn[{%
            \renewcommand\twocolumn[1][]{#1}%
            \maketitle
            \begin{center}
    \centering
    \captionsetup{type=figure}
    \includegraphics[width=\textwidth]{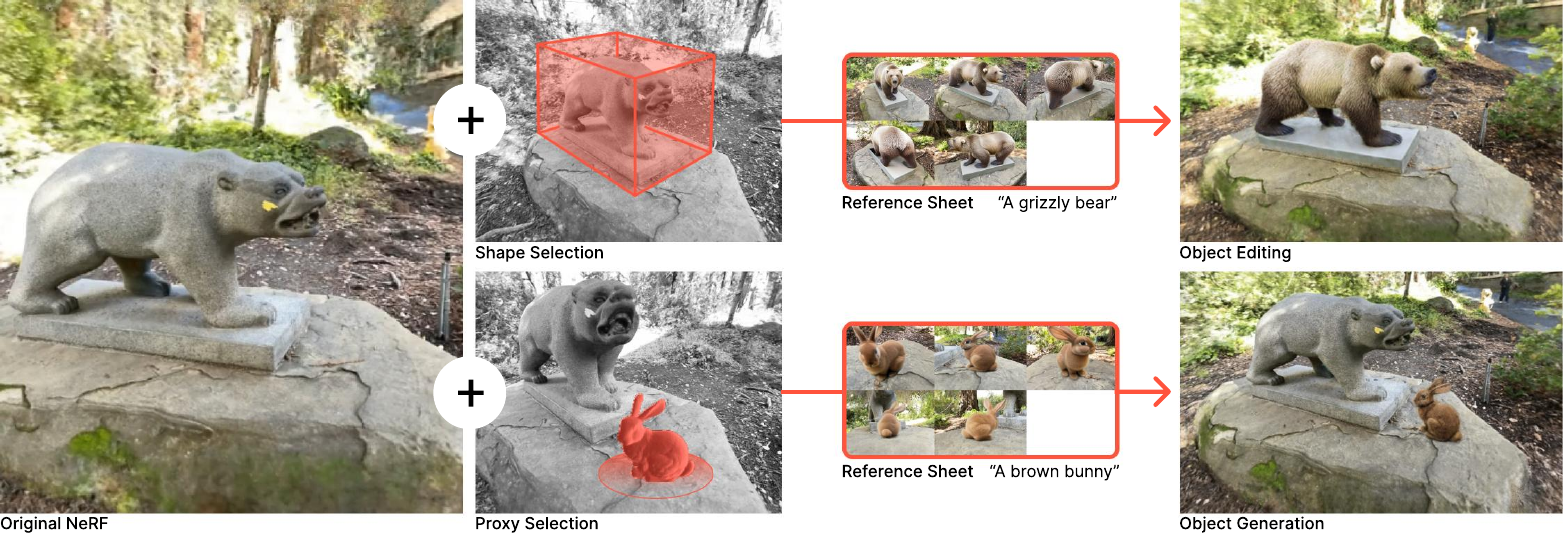}
    \caption{\textbf{SIGNeRF} -- A method that enables NeRF scene editing by either shape selection for object modification or by proxy object placement to insert a synthesized object. The transformed scene is represented by a NeRF trained on a set of edited input images, which are efficiently generated by a diffusion process using a multiview consistent reference sheet. Project page  available at \href{https://signerf.jdihlmann.com/}{\textcolor{blue}{https://signerf.jdihlmann.com/}}
    }
    \label{fig:introduction:overview}
\end{center}

        }]

\begin{abstract}
    Advances in image diffusion models have recently led to notable improvements in the generation of high-quality images.
    In combination with Neural Radiance Fields (NeRFs), they enabled new opportunities in 3D generation.
    However, most generative 3D approaches are object-centric and applying them to editing existing photorealistic scenes is not trivial.
    We propose SIGNeRF, a novel approach for fast and controllable NeRF scene editing and scene-integrated object generation.
    A new generative update strategy ensures 3D consistency across the edited images, without requiring iterative optimization.
    We find that depth-conditioned diffusion models inherently possess the capability to generate 3D consistent views by requesting a grid of images instead of single views.
    Based on these insights, we introduce a multi-view reference sheet of modified images.
    Our method updates an image collection consistently based on the reference sheet and refines the original NeRF with the newly generated image set in one go.
    By exploiting the depth conditioning mechanism of the image diffusion model, we gain fine control over the spatial location of the edit and enforce shape guidance by a selected region or an external mesh. \vspace{-0.8\baselineskip}

\end{abstract}

\section{Introduction}
\label{sec:intro}

In this work, we focus on scene-integrated generation for editing existing 3D scenes using generative 2D diffusion models~(Fig.~\ref{fig:introduction:overview}). We propose an approach to create or modify objects within existing NeRF~\cite{NeRF} scenes while preserving the original scenes' structure and appearance.
Previous methods such as Instruct-NeRF2NeRF~\cite{Instruct-NeRF2NeRF} and DreamEditor~\cite{DreamEditor} already demonstrated scene modification with text instructions.
However, their pipelines are complex, involve additional training and lack fine-grained control.

We observe that ControlNet~\cite{ControlNet}, an image diffusion model with additional depth guidance is inherently capable of generating coherent and consistent views of an object when applied to image tiles in a grid layout.
Using this combined image generation, we construct a \textit{reference sheet} of edits that can subsequently be applied to the original NeRF scene by updating the underlying image dataset. The generation process of the reference sheet is conditioned on the existing NeRF scene and a text prompt that can be tuned easily to one's liking.
This editing capability is further refined by introducing two selection methods in scene space to condition the edits of the diffusion model. A mesh proxy can be placed and composed with the depth data of the existing NeRF to guide the insertion of new objects or a rough bounding box can be used to select existing parts of the scene.
This addition grants enhanced control over the position and visual representation of generated entities and preserves the quality of the unedited parts.
In summary, our method consists of the following key elements:
\begin{itemize}
    \item Reference-sheet-based assembly as a simple and easy-to-implement approach to improve consistency in generated multi-view image data when using the ControlNet~\cite{ControlNet} image diffusion model.
    \item A modular pipeline to update a NeRF dataset based on a reference sheet that is generated and refined to the user's liking using text prompts. This update scheme can be easily repeated to further increase quality if needed.
    \item Fine-grained control of the generation process by offering multiple selection modes to constrain edits in scene space.
\end{itemize}

\noindent Compared to existing methods like DreamEditor~\cite{DreamEditor} and Instruct-NeRF2NeRF~\cite{Instruct-NeRF2NeRF} we introduce a simplified, more modular pipeline that achieves comparable or improved results while adding control to the editing process.

\begin{figure*}[t!]
    \centering
    \includegraphics[width=\textwidth]{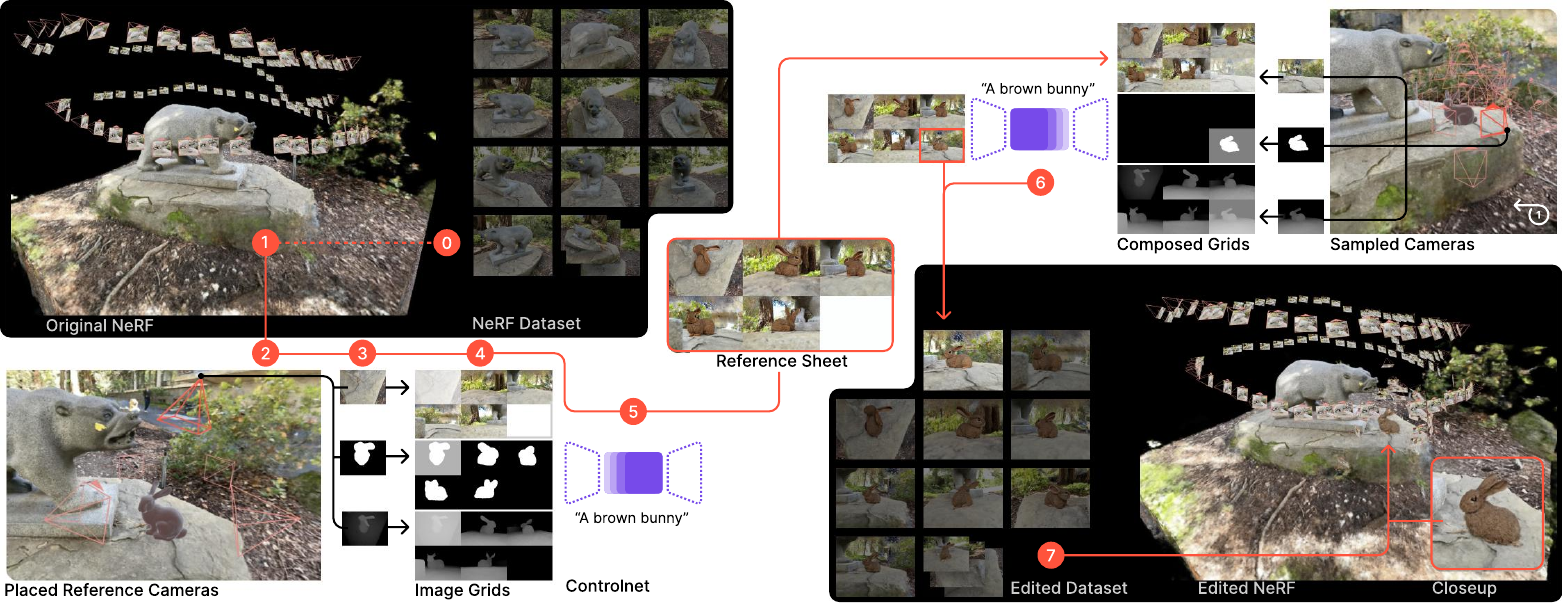}
    \caption{\textbf{SIGNeRF pipeline for NeRF scene editing} -- here, object generation. First, the original NeRF scene is trained \textbf{(1)}, and a proxy object is placed into the scene \textbf{(2)}. After a precise selection, we place reference cameras ($5$ here) into the scene \textbf{(3)}, render the corresponding color, depth, and mask images, and arrange them into image grids \textbf{(4)}. These grids are used to generate the reference sheet with conditioned image diffusion \textbf{(5)}. To propagate the edits to the entire image set, for each camera, a color, depth, and mask image are rendered and placed into the empty slot of the fixed reference sheet.
        We generate a new edited image consistent with the reference sheet by leveraging an inpainting mask. The step is repeated for all cameras \textbf{(6)}.
        Finally, the NeRF is fine-tuned on the edited images \textbf{(7)}}
    \label{fig:methods:pipeline}
\end{figure*}

\section{Related Work}
\label{sec:related}

\begin{paragraph}{Text-to-Image Generation}
    \label{pg:related_work:text-to-image_generation}
    Recent advances in diffusion probabilistic models~\cite{DiffusionProbalisticModels, AdvancesInDPM} in combination with the availability of large datasets~\cite{LAION}, enabled the generation of high-resolution and diverse images~\cite{GLIDE, StableDiffusion, Imagen}.
    Fine-tuning these models can significantly enhance the capacity to guide and personalize image generation~\cite{LoRA, DreamBooth}.
    In addition, ControlNet~\cite{ControlNet} has demonstrated that image diffusion models can be fine-tuned to condition them on various secondary inputs, such as depth, poses, or even sketches~\cite{SketchGuidance}.
    Furthermore, editing existing images in a generative manner with instructions is possible as outlined by InstructPix2Pix~\cite{InstructPix2Pix}.
\end{paragraph}

\begin{paragraph}{Text-to-3D Generation}
    \label{pg:related_work:text-to-3d_generation}
    The field of text-to-3D generation has recently experienced a significant breakthrough, with the combination of image diffusion models and NeRFs~\cite{DreamFusion}.
    Previously, the emphasis of text-driven 3D generation relied on optimizing meshes~\cite{Text2Mesh, CLIP-Mesh, TANGO}, point clouds~\cite{CLIP-Forge}, voxels~\cite{Text2Shape}, and NeRFs~\cite{DreamFields}, guided by Contrastive Language-Image Pretraining (CLIP)~\cite{CLIP} embeddings and supervision.
    While mesh generation guided by CLIP continues to be an active area of research~\cite{GET3D}, recent advances in image diffusion models address a significant challenge in text-to-3D generation, namely the scarcity of 3D data.
    The pioneering work by DreamFusion~\cite{DreamFusion} introduced Score-Distillation-Sampling~(SDS), leveraging a pre-trained image diffusion model conditioned on a text input prompt to generate novel 3D objects.
    Building upon this, Latent-NeRF~\cite{LatentNeRF} extended DreamFusion to the latent domain, accelerating the generation process using Stable Diffusion~\cite{StableDiffusion}.

    The generation quality and object variation can be further improved by modifying SDS~\cite{Magic3D, ProlificDreamer, DreamTime}.
    Although SDS has proven itself as a powerful tool for 3D object generation, it requires a large amount of memory and time to train.
    In contrast, Instruct-NeRF2NeRF~\cite{Instruct-NeRF2NeRF} proposed that conditioned image diffusion models can be used for 3D object generation without requiring a direct gradient update, making it vastly faster.
    By iteratively updating the NeRF image set with conditioned generated images, the NeRF can be trained to generate the desired 3D object.
    The final appearance, however, emerges slowly over time while our method gives an immediate preview.

    While there are other promising paths to 3D generation, including the use of 3D diffusion models~\cite{DiffRF, Diffusion-SDF, Point-E} and synthesizing novel views from single or multiple images to lift an image representation into 3D space~\cite{3DiM, NeuralLift-360, NeRDi, SparseFusion, Zero-1-to-3, Make-It-3D, One-2-3-45}, this work utilizes 2D image diffusion, given its applicability to NeRF representations.
\end{paragraph}

\begin{paragraph}{NeRF Editing}
    \label{pg:related_work:nerf_editing}
    Although NeRFs have proven themselves as a state-of-the-art tool for novel view synthesis and 3D scene reconstruction~\cite{NeRF, Mip-NeRF360, InstantNGP}, editing NeRFs is still a challenging task and is the subject of ongoing research.
    Efforts to manipulate NeRFs were initiated by NeRF-Editing~\cite{NeRF-Editing}, which proposed a method to deform NeRFs by modifying the underlying implicit function.
    NeRFShop~\cite{NeRFShop} extended the work of cage-based deformation~\cite{DeformingNeRF}, introducing an intuitive selection of objects for both affine and non-affine transformations and object duplication.
    However, these approaches are limited to simple deformations and do not allow for more complex edits or additions to the scene.
    In contrast, NeuMesh~\cite{NeuMesh} learns a disentangled neural
    mesh-based implicit field to edit geometry and appearance, enabling geometry deformation and texture swapping, filling, and painting.
    It has also been demonstrated that NeRFs can be combined by inserting objects into pre-existing NeRF scenes~\cite{Control-NeRF}.
    Despite the strides made in NeRF editing, the existing solutions provide only a basic level of editing functionality compared to conventional 3D editing software.
    Generally, obtaining visually pleasing results still requires artistic proficiency and often manual labor.
\end{paragraph}

\begin{paragraph}{Generative NeRF Editing}
    \label{pg:related_work:generative_nerf_editing}
    With advances in text-to-3D generation, a new area in NeRF editing emerged, incorporating generative 3D models to modify existing NeRF scenes.
    Set-the-Scene~\cite{Set-the-Scene} and Compositional 3D~\cite{Compositional3D} present methods for controlled scene generation using proxy objects and bounding boxes. In addition, Composable 3D Diffusion enables moving generated objects within the composed scene, which consists of multiple NeRFs.
    On the other hand, SINE~\cite{SINE} allows direct NeRF editing by changing a reference image in 2D space.
    It uses an editing field to adjust the template NeRF's geometry and appearance to match the image changes.
    Similarly, VOX-E~\cite{VOX-E} combines an original NeRF with an edited NeRF generated using text-based SDS.
    It merges the field by using the attention mask from the image diffusion model.
    Yet, these methods often result in overlaying edits and increased memory and training time due to using multiple NeRFs.
    In contrast, Instruct-NeRF2NeRF~\cite{Instruct-NeRF2NeRF} uses an Iterative Dataset Update~(IDU) strategy to edit NeRF's image dataset with InstructPix2Pix~\cite{InstructPix2Pix}, such that the NeRF can be transformed based on editing instructions.
    Inspired by the work of Instruct-NeRF2NeRF~\cite{Instruct-NeRF2NeRF}, we also update the NeRF image dataset with an image diffusion model.
    However, unlike Instruct-NeRF2NeRF, we do this in a preprocessing stage using the depth-conditioned ControlNet~\cite{ControlNet}, gaining more control over the generational process.
    DreamEditor~\cite{DreamEditor} also allows for controlled text-based scene editing by focusing the selection with finetuning an image diffusion model DreamBooth~\cite{DreamBooth} before using SDS.
    Nevertheless, using SDS and DreamBooth is very time-consuming compared to our approach.
    Furthermore, Instruct-NeRF2NeRF and DreamEditor can be challenging to control by only semantics and often fail to generate new objects within the scene at specific locations.
    Blended-NeRF~\cite{Blended-NeRF} improves positional control by limiting the generation within manually controlled bounding boxes.
    However, Blended-NeRF results often do not fit the scene and get clipped by its bounding boxes, highlighting the need for further improvements in controlled generative NeRF editing.
    We introduce a new approach to generative NeRF editing, providing more precise control over the generated edits with a manual selection or proxy object guidance to generate new and fitting objects within an existing scene.
\end{paragraph}

\begin{figure*}[t!]
    \centering
    \includegraphics[width=\textwidth]{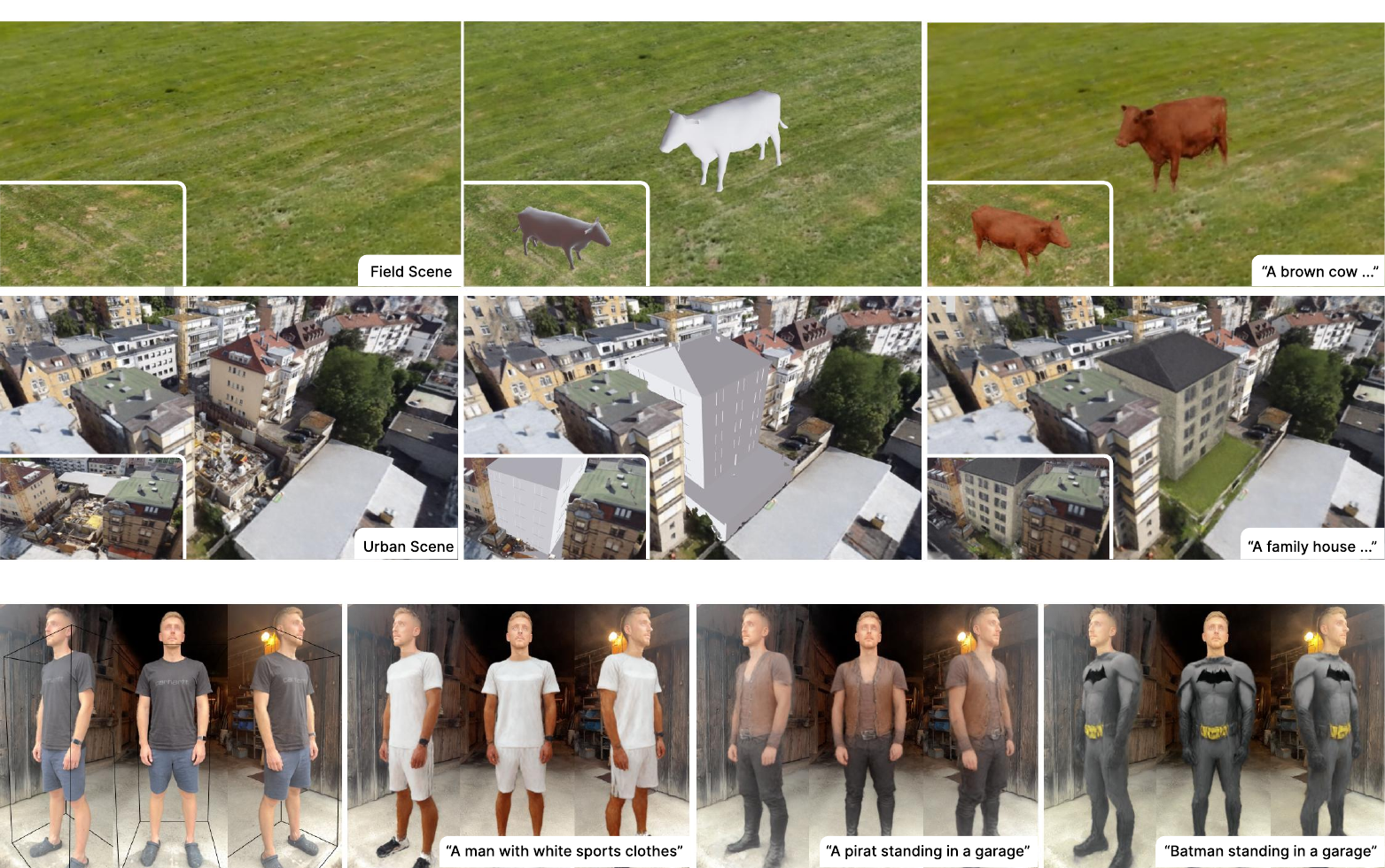}
    \caption{\textbf{Object insertion and object modification.} -- (top) The cow geometry is centrally placed on a meadow to obtain a photorealistic scene. (middle) Note how occlusions are properly handled when generating the synthetic house based on the inserted proxy. (bottom) Objects can easily be transformed based on a prompt. Due to the more complex surface texture and geometric changes the pirate and the Batman costume required an additional iteration to obtain the same level of consistency compared to the simpler sports clothes.
    }
    \label{fig:experiments:qualitative_results}
\end{figure*}

\section{Method}
\label{sec:method}

\begin{figure}[t!] %
  \centering
  \includegraphics[width=\linewidth]{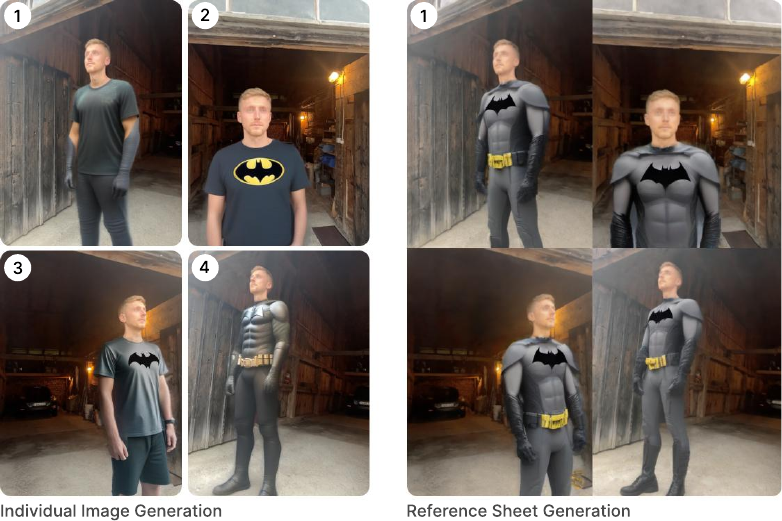}
  \caption{\textbf{Reference Sheet Generation} -- Using ControlNet \cite{ControlNet} inpainting to edit scene parts image-by-image results in drastically different looks per view (left) although all parameters and the seed are the same. In contrast, we obtain a consistent reference sheet (right) by arranging the input images into a grid, letting ControlNet process the entire sheet in a single generation step.
  }
  \label{fig:method:reference_sheet_generation}
\end{figure}

SIGNeRF is a method for scene-integrated generation, including edits and object generation within an existing NeRF scene.
At the core, we introduce the concept of a reference sheet image grid to maintain multi-view coherence and to gain control over the generation process~(Figs.~\ref{fig:methods:pipeline} and ~\ref{fig:method:reference_sheet_generation}).
Given a NeRF model trained on the original image set, a set of views is selected to compose a grid of images.
The scene edits are performed on this grid by ControlNet, a conditioned image diffusion model~(Sec.~\ref{chpt:background:image_generation:task_specific_diffusion:controlnet}) in one go to produce the reference sheet. In the second step, the reference sheet constrains the generation of an updated version of the full image set.
We observe that this two-step procedure already supports generating quite consistent edited views.
The edited 3D scene is obtained by finetuning the original NeRF scene with these newly generated views.
Optionally, if the multi-view consistency needs to be improved further a second iteration can be performed, where the reference sheet is updated based on the once-edited NeRF to generate a second image set.
An overview of the pipeline on consistent grid image generation and subsequent scene-integrated generation is presented in Fig.~\ref{fig:methods:pipeline}.

\begin{subsection}{Background}

    \begin{paragraph}{NeRF}
        Neural Radiance Fields (NeRFs)~\cite{NeRF} implicitly represent a scene by learning a continuous function of volumetric density and color.
        A 5D coordinate, composed of a spatial location~$(x,y,z)$ and a viewing direction~$(\theta, \varphi)$, is mapped to a view-dependent emitted radiance~$(r,g,b)$ and a volume density~$\sigma$.
        Volumetric rendering accumulates the densities and colors at multiple sample locations along the view ray ~$\textbf{r}_j$ of a virtual camera $C$ to obtain the final pixel color~$\hat{C}(\textbf{r}_j)$ to either calculate a novel view or to update the neural representation based on the difference to some of the input images ~$\textbf{I} = \{\textbf{I}_1, ..., \textbf{I}_N\}$.
    \end{paragraph}

    \begin{paragraph}{ControlNet}
        \label{chpt:background:image_generation:task_specific_diffusion:controlnet}
        ControlNet~\cite{ControlNet} is a specific image diffusion model that allows for constraining the image generation process with additional conditions, such as sketches, edge, or depth maps.
        In the typical process of image-to-image diffusion, an image $x$ is first encoded with an encoder~$\mathcal{E} (x) = z$ to produce a latent image $z$.
        Then the latent image $z$ is iteratively updated by the diffusion model, which is guided by a noise predictor UNet~$\epsilon_{\theta}$ conditioned on the encoded text input~$\tau_{\theta}(\boldsymbol{y})$ and the current time step~$t$.
        The final latent image $z$ is then decoded by a decoder~$\mathcal{D}$ to obtain the generated image~$\boldsymbol{y}$.
        In the case of ControlNet, the image generation process is guided by an additional condition~$\boldsymbol{c_f}$, that serves as an additional input to the noise predictor UNet~$\epsilon_{\theta}$, (Eq.~\ref{eq:methods:controlnet}).
        \begin{equation}
            \label{eq:methods:controlnet}
            \epsilon = \epsilon_{\theta} \left( z_t,t,\boldsymbol{c_f},\tau_{\theta}(\boldsymbol{y}) \right)
        \end{equation}
        This functionality is used in our pipeline to guide the image-to-image generation process with depth maps.
    \end{paragraph}

\end{subsection}

\begin{subsection}{Controlled Consistent Generation}
    The key challenge of 3D generation techniques is to generate consistent views with an image diffusion model. Our approach is based on reference sheet generation which is simpler, faster and features direct control compared to the methods introduced in Sec.~\ref{sec:related} which rely on a repetitive cycle of intertwined diffusion and NeRF updates.

    \begin{paragraph}{Reference Sheet Generation} We observe that the image diffusion model ControlNet~\cite{ControlNet} can already generate multi-view consistent images of a scene without the need for iterative refinement.
        While generating individual views sequentially introduces too much variation to integrate them into a consistent 3D model, arranging them in a grid of images that are processed by ControlNet in one pass significantly improves the multi-view consistency as depicted in Fig.~\ref{fig:method:reference_sheet_generation}.
        Based on the depth maps rendered from the original NeRF scene we employ a depth-conditioned inpainting variant of ControlNet to generate such a reference sheet of the edited scene.
        A mask specifies the scene region where the generation should occur.

        This step gives a lot of control to the user. Different appearances can be produced by generating reference sheets with different seeds or prompts. The one sheet finally selected will directly determine the look of the final 3D scene.
    \end{paragraph}

    \begin{paragraph}{Image Set Update} Despite the potential of grid generation, we are limited in the number of images we can place in one sheet due to the memory and attention limitations of the image diffusion model.
        Given that a proper NeRF training might require hundreds of images, an alternative method for generating consistent scene views is necessary.
        We solve this challenge in a two-step process.
        First, we generate the reference sheet and subsequently use it to iteratively update the images in the NeRF dataset image-by-image.
        To ensure that all other views are also consistent with the reference sheet, we originally left one slot empty when producing the reference sheet, e.g. the bottom right corner, and condition this slot with the appropriate original image, mask and depth of the to-be-generated view while keeping the rest of the reference sheet fixed. This way, the appearance of the reference sheet is propagated to all input views.

    \end{paragraph}
\end{subsection}

\begin{figure*}[t!]
    \centering
    \includegraphics[width=\textwidth]{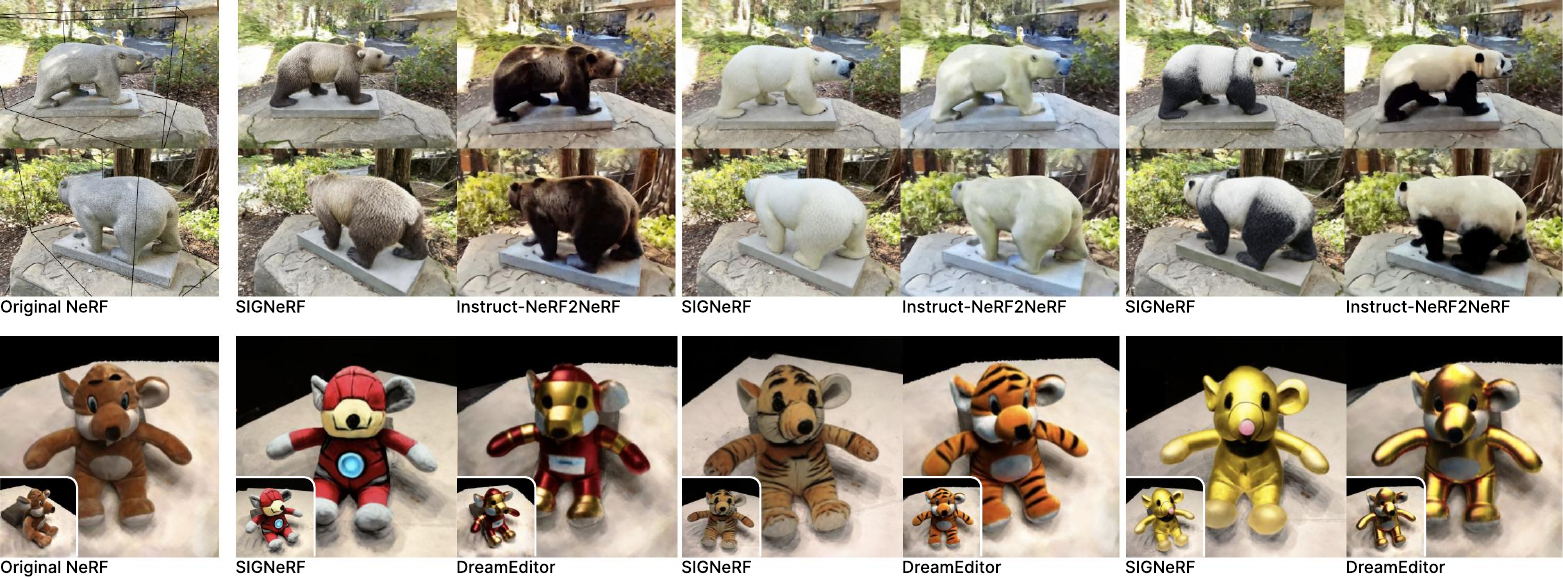}
    \caption{\textbf{Qualitative Comparison} -- SIGNeRF results are compared to Instruct-NeRF2NeRF~\cite{Instruct-NeRF2NeRF} (top) and DreamEditor~\cite{DreamEditor} (bottom). For the bear, the generated fur texture with SIGNeRF (left) shows a more distinguished structure and the snout regions is clearly more consistent. Compared to DreamEditor the images are different but the image quality comparable. }
    \label{fig:experiments:comparison}
\end{figure*}

\begin{subsection}{Scene Integrated Generation}
    \label{chpt:methods:signerf}
    \label{sec:methods:signerf:scene_integrated_generation}

    Based on these ideas we present the full scene-integrated generation pipeline as depicted in Fig.~\ref{fig:methods:pipeline} for the specific case of object generation.
    \begin{enumerate}
        \item \textbf{Original NeRF Scene} -- A NeRF scene~$\boldsymbol{S}$ is reconstructed using set of $N$ input images~$\boldsymbol{I}_S = \{I^S_1, I^S_2, ..., I^S_N\}$ alongside the corresponding cameras~$\boldsymbol{C}_S = \{C^S_1, C^S_2, ..., C^S_N\}$. This NeRF serves as the foundation for the subsequent scene-integrated generation.
        \item \textbf{Object Selection} -- The next step is to mark the 3D region to be edited. For region-based edits, we use a bounding box. To introduce a new object, we use a proxy mesh that is placed in the scene.
        \item \textbf{Reference Camera Placement} -- Around the selected edit region, we place the $M << N$ cameras~$\boldsymbol{C}_R = \{C^R_1, C^R_2, ..., C^R_M\}$ for the reference sheet generation. They need to cover a sufficient range of views around the object and have the object properly centered.
        \item \textbf{Reference Input Image Rendering} -- For each reference camera we use the original NeRF scene~$\boldsymbol{S}$ to render the RGB images
              ~$\boldsymbol{I}_R = \{I^R_1, I^R_2, ..., I^R_M\}$, with corresponding depths~$\boldsymbol{D}_R$ and inpainting masks~$\boldsymbol{M}_R$.
        \item \textbf{Reference Sheet Assembly and Generation} -- Each set of $M$ reference input images is arranged in a 2D grid resulting in ~$\bar{\boldsymbol{I}}_R$, ~$\bar{\boldsymbol{D}}_R$ and ~$\bar{\boldsymbol{M}}_R$, respectively.
              One grid cell remains empty on purpose.
              The reference sheet~$\boldsymbol{R}$ is generated by piping the created grids as inputs and conditions to ControlNet with the selected prompt~$y$:
              \begin{equation}
                  \boldsymbol{R} \leftarrow \text{ControlNet}(\bar{\boldsymbol{I}}_R, \bar{\boldsymbol{D}}_R, \bar{\boldsymbol{M}}_R,  y)
              \end{equation}
        \item \textbf{Image Set Update} -- After producing the desired reference sheet, it is used to generate a new image dataset~$\boldsymbol{I}_{\hat{S}} = \{I^{\hat{S}}_1, I^{\hat{S}}_2, ..., I^{\hat{S}}_N\}$. We first create depth maps~$\boldsymbol{D}_S = \{D^S_1, D^S_2, ..., D^S_N\}$ and inpainting masks~$\boldsymbol{M}_S = \{M^S_1, M^S_2, ..., M^S_N\}$ from the original NeRF scene~$\boldsymbol{S}$ and then replace the empty grid cells with the corresponding image or depth map. This results in the following update rule for the image dataset~$\boldsymbol{I}_{\hat{S}}$:
              \begin{equation}
                  \begin{split}
                      I^{\hat{S}}_i \leftarrow & \text{ControlNet}(\boldsymbol{R}_i, \bar{\boldsymbol{D}}^R_i, \boldsymbol{M}^S_i,  y) \quad \forall i \in \{1, 2, ..., N\}                                         \\
                                               & \text{with } \boldsymbol{R}_i \leftarrow \boldsymbol{R} \oplus \{I^S_i\}  \text{ and } \bar{\boldsymbol{D}}^R_i \leftarrow \bar{\boldsymbol{D}}_R \oplus \{D^S_i\}
                  \end{split}
              \end{equation}
        \item \textbf{Finetuning the NeRF Scene} -- The final step is to finetune the original NeRF scene~$\boldsymbol{S}$ with the generated image dataset~$\boldsymbol{I}_{\hat{S}}$ to receive the edited scene~$\boldsymbol{\hat{S}}$.
        \item \textbf{Multiple Iteration (optional)} -- While the generated reference sheet always shows a consistent style in all tiles it might happen that the underlying 3D shape is not yet fully consistent. If this leads to visible artifacts Steps 4 to 7 can be optionally repeated once more, this time starting with the updated NeRF~$\boldsymbol{\hat{S}}$ to render the reference input images. The parameters of the ControlNet in Steps 5 and 6 have to be tuned to stick closer to the input in this second iteration.
    \end{enumerate}
\end{subsection}

\noindent A key feature of our pipeline is its modularity.
Each step can be developed and optimized independently. Compared to the tight iterative NeRF/image diffusion update loops of other approaches, in our pipeline individual steps can easily be exchanged, e.g.\ to enable different scene modifications, or to repeat some as indicated in the optional Step 8.

\begin{paragraph}{Selection Modes}
    \label{sec:method:selection}
    For precise control over the generation location, we introduce two basic selection modes: Shape selection and proxy selection~(Fig.~\ref{fig:introduction:overview}).
    With shape selection, a region of the scene can be selected by an axis-aligned bounding box.
    We use this bounding box and generate a per-camera mask by comparing its depth to the rendered NeRF depth.
    Further, we a combined depth map by clamping the rendered NeRF depth within the point closest and furthest away to the rendering camera within the bounding box.

    With the proxy selection mode, one can position an arbitrary mesh within the NeRF scene.
    Similar to the shape selection, we generate a depth map and a mask for each camera view, but herefore combine the rendered depth map of the NeRF scene with the depth map rendered of the proxy object relative to the camera.
    In difference, we use the visible part of the proxy object in the rendered view as the mask.
    As previously described the generated depth map conditions the image diffusion model, while the generated mask is used as an inpainting mask, to blend the generated image with the original NeRF image.
    The masks can further be dilated in image space to allow control over the to-be-generated area.
\end{paragraph}

\begin{paragraph}{Reference Sheet}
    The quality of the reference sheet directly impacts the generation results.
    One important aspect is the number of reference cameras and their position within the scene.
    Optimally, we use the fewest cameras necessary to capture the region of interest from all angles.
    This strategy minimizes generation time while maintaining consistency.
    Another criterion is the proximity of reference cameras to the original cameras.
    Generally, it is best to render reference views from a position close to the original camera positions.
    In cases where the edit region is too distant from any original camera, the reference cameras need to move closer in order to increase the number of pixels in the masks for more details in the generated edits.
    A standout feature of SIGNeRF is its ability to preview the reference sheet before generating the complete updated image set. Besides choosing the intended appearance we recommend iterative adjustments to the reference sheet by replacing undesired generated images of the image grid until satisfactory results are obtained.
\end{paragraph}

\section{Experiments}
The proposed pipeline facilitates the generation of new objects within a scene and the editing of existing ones.
We assess the quality of the 3D scenes generated by SIGNeRF and compare it to existing methods.

\begin{subsection}{Experimental Setup}
    \begin{paragraph}{Datasets}
        We use various types of scenes for our experiments, most of them in real-world settings. While some are front-facing scenes, our primary focus is on~$360$° view scenes due to their inherent challenges and potential in 3D~scene generation.
        Datasets from Instruct-NeRF2NeRF~\cite{Instruct-NeRF2NeRF}, DTU~\cite{DTU} and BlendedMVS~\cite{BlendedMVS} are utilized.
        Additionally, to address the scarcity of realistic NeRF datasets, we created custom scenes using smartphones with PolyCam~\cite{Polycam} or drones.
        Camera parameters were either sourced directly from relevant apps or inferred using COLMAP~\cite{Colmap}.
        The scenes contain $30$ to~$300$ images.
    \end{paragraph}

    \begin{paragraph}{Implementation Details}
        SIGNeRF is built upon Nerfstudio using Nerfacto~\cite{Nerfstudio} as the underlying NeRF implementation.
        For the image diffusion model, we modified an inpainting version of ControlNet~\cite{ControlNet, StableDiffusion, ControlNetInpainting, SDXL} with the SDXL~\cite{SDXL}-diffusion backbone to allow for conditioning on masked content.
        ControlNet scale is set between~$[0.4, 1.0]$, guidance values range from~$[6, 10]$ and denoising strength varies between~$[0.5, 0.95]$.
        However, these parameters can be adjusted as needed.
    \end{paragraph}

\end{subsection}

\begin{table}
    \centering
    \renewcommand\tabularxcolumn[1]{m{#1}}
    \begin{tabularx}{\linewidth}{
            l
            S[table-format=1.4,table-column-width=0.115\linewidth]
            S[table-format=1.4,table-column-width=0.115\linewidth]
            S[table-format=1.4,table-column-width=0.115\linewidth]
            S[table-format=1.4,table-column-width=0.115\linewidth]
        }
        \toprule
        \multirow{2}{*}[-1.5em]{\thead{Method}} & {\thead{Image}}                                           & {\thead{NeRF}}            & \multicolumn{2}{c}{\thead{Background Preserv.}}                 \\
        \cmidrule(r){2-3} \cmidrule(r){4-5}     & \multicolumn{2}{c}{\thead{CLIP T2I Dir. Sim.~$\uparrow$}} & {\thead{PSNR $\uparrow$}} & {\thead{SSIM~$\uparrow$}}                                       \\
        \midrule
        Instruct-N2N \cite{Instruct-NeRF2NeRF}  & 0.1603                                                    & 0.1600                    & \text{30.09}                                    & \text{0.64}   \\
        DreamEditor \cite{DreamEditor}          & \text{-}                                                  & 0.1849                    & \text{-}                                        & \text{-}      \\
        Ours                                    & \textbf{0.23}                                             & \textbf{0.2125}           & \textbf{32.22}                                  & \textbf{0.83} \\
        \bottomrule
    \end{tabularx}
    \caption{\textbf{Quantitative Evaluation} -- Highlighting the CLIP text-to-image directional similarity comparing the diffusion \textit{Image} edit (rendered NeRF view edited with the method-specific diffusion mode) and the trained \textit{NeRF} render (rendered view from the edited NeRF) to the original image with corresponding prompts. The background preservation capability of our method is evaluated using PSNR and SSIM by masking the edited object, comparing the original and the edited NeRF result.}
    \label{tab:experiments:quantiative_comparison}
\end{table}

\begin{subsection}{Qualitative Evaluation}

    Fig.~\ref{fig:experiments:qualitative_results} demonstrates SIGNeRF's object generation and editing capabilities (see also the videos in the supplemental material).
    In the first two rows, novel objects are inserted with fine-grained control of position, orientation and size. The objects are synthesized, and conditioned on a geometry proxy. These objects are integrated seamlessly into the scene, exhibiting fitting lighting and texture properties. Notably in the cow scene, one does not even need strong trackable features for precise placement.

    In the last row, an existing object is modified based on a text prompt to generate different appearances. Note that the desired edits only affect the marked object. SIGNeRF is, however, powerful enough to even adjust the geometry where necessary, e.g. short vs.\ long trousers, while the background is preserved.

\end{subsection}

\begin{figure}[t!] %
    \centering
    \includegraphics[width=\linewidth]{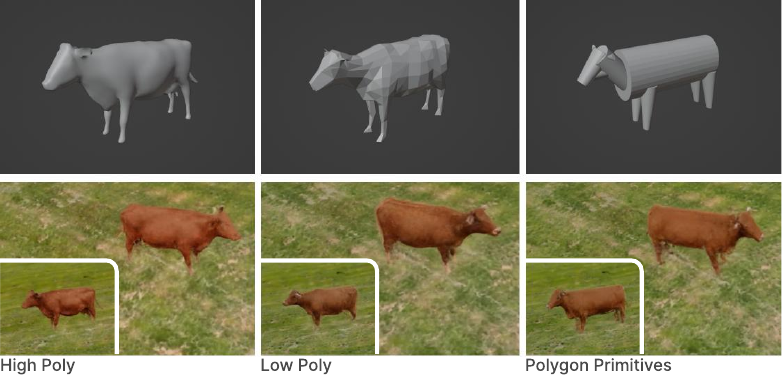}
    \caption{\textbf{Influence of the proxy geometry} -- The synthetic cow is generated with three different proxy meshes with the prompt "A brown cow". From left to right: High-poly proxy mesh, low-poly proxy mesh, and simple geometric primitives. }
    \label{fig:experiments:abblation}
\end{figure}

\begin{paragraph}{Proxy Shape}
    The impact of the proxy shape and geometric detail is visualized in ~Fig.~\ref{fig:experiments:abblation}.
    The shape of the proxy mesh does influence the final object as ControlNet tries to fit the depth condition created with the proxy mesh, as apparent when the body of the cow is approximated by a cylinder. Nevertheless, the results indicate that the proxy mesh does not need to be a detailed representation of the object, as the diffusion process added additional geometry details. Both low and high-poly versions yield satisfactory results.
\end{paragraph}
\begin{paragraph}{ControlNet Guidance}
    Further, we studied the impact of the scale parameter of ControlNet~\cite{ControlNet}, which handles the closeness of the generation to the depth condition, on the final NeRF edit.
    We observe that a low scale strongly reduces the impact of the condition, leading to an edit that deviates largely from the original shape.
    This allows us to make drastic changes, like the Batman example~(Fig.~\ref{fig:method:reference_sheet_generation}, Fig.~\ref{fig:experiments:qualitative_results}).
    However, a low scale can also lead to more irritation in the 3D consistency of the generated image, producing artifacts in the NeRF.
    In these cases, a second iteration of our process~(Sec.~\ref{sec:methods:signerf:scene_integrated_generation}),  using a low scale in the first and a high scale in the second iterator achieves the desired results.
    In~Fig.~\ref{fig:experiments:qualitative_results}, the Pirate and the Batman are generated with a second generation iteration, while the white shirt is generated directly.
\end{paragraph}

\begin{figure}[t!] %
    \centering
    \includegraphics[width=\linewidth]{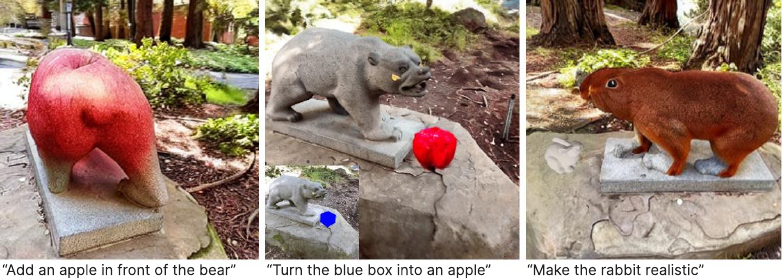}
    \caption{\textbf{Instruct-NeRF2NeRF} -- Instruct-NeRF2NeRF fails to generate new objects in the scene, even when merging proxy meshes with the NeRF scene. Compare to our results in Fig.~\ref{fig:introduction:overview}}
    \label{fig:experiments:instructnerf2nerf_failing}
\end{figure}

\begin{subsection}{Comparison}
    Scene editing results of SIGNeRF are compared to Instruct-NeRF2NeRF~\cite{Instruct-NeRF2NeRF} and DreamEditor~\cite{DreamEditor} in Fig.~\ref{fig:experiments:comparison}. While Instruct-NeRF2NeRF~\cite{Instruct-NeRF2NeRF} produces washed-out textures and suffers from the Janus effect of showing different faces from different views for the bear, our generated results show a more consistent snout region and more vivid, more structured pelt textures.
    Here, SIGNeRF is superior in terms of scene preservation, selection precision, generation quality, and color integrity.

    In the second row of Fig.~\ref{fig:experiments:comparison}, it achieves similar results to DreamEditor, while both methods have some artifacts that cannot be directly compared.
    DreamEditior tends to generate simpler and over-smoothed objects due to the high classifier-free guidance needed for score-destillation-sampling~\cite{ProlificDreamer}.
    In contrast, SIGNeRF generates more complex and realistic-looking objects but may show consistency artifacts for highly detailed regions.

    Both other methods are designed primarily for scene editing, generating new objects semantically independent from the scene is not possible, let alone controlling the object's position, scale, rotation, and shape within the scene.
    For example, Instruct-NeRF2NeRF fails to generate a rabbit in front of the bear statue (Fig.~\ref{fig:experiments:instructnerf2nerf_failing}) while SIGNerF embeds it consistently into the scene (Fig.~\ref{fig:introduction:overview}).
    But also for editing tasks, SIGNeRF is superior as it also allows semantically dependent parts to be unedited, e.g. the face in the Batman scene.

    Another aspect is the time spent in the generation process.
    Instruct-NeRF2NeRF and DreamEditor, require more than an hour for a generation that is only fully visible at the end.
    While SIGNeRF only takes half the time on a single GPU, using one dataset iteration, it additionally provides a preview option with the reference sheet, allowing the user to adjust the output until satisfied before starting the generation process. Furthermore, the image set update (Step 6) can easily be parallelized over all images in our approach while the interlocked NeRF/image updates in the other methods are purely sequential.
\end{subsection}

\begin{subsection}{Quantitative Evaluation}
    \label{sec:experiments:quantitative}
    Even though the process of generating and editing 3D scenes is inherently subjective, in line with previous works we utilize the CLIP~\cite{CLIP} text-to-image directional similarity to provide a quantitative perspective.
    This metric~\cite{StyleGANNADA} evaluates the semantic distance between the original image and edited NeRF scene to their corresponding prompt pairs.
    Table~\ref{tab:experiments:quantiative_comparison} compares the scores averaged over a total of~$10$~scenes to  Instruct-NeRF2NeRF~\cite{Instruct-NeRF2NeRF} and DreamEditor~\cite{DreamEditor}

    Another metric provided in Table~\ref{tab:experiments:quantiative_comparison} accounts for background preservation calculated by masking the edited object and comparing the original and edited NeRF background.
    Due to their compressive nature image diffusion models are prone to degrading the image quality.
    Since SIGNeRF employs a masked update strategy, the background is better preserved compared to Instruct-NeRF2NeRF, which uses a repeated iterative update strategy.

\end{subsection}

\begin{subsection}{Limitations}
    \begin{figure}[t!] %
    \centering
    \includegraphics[width=\linewidth]{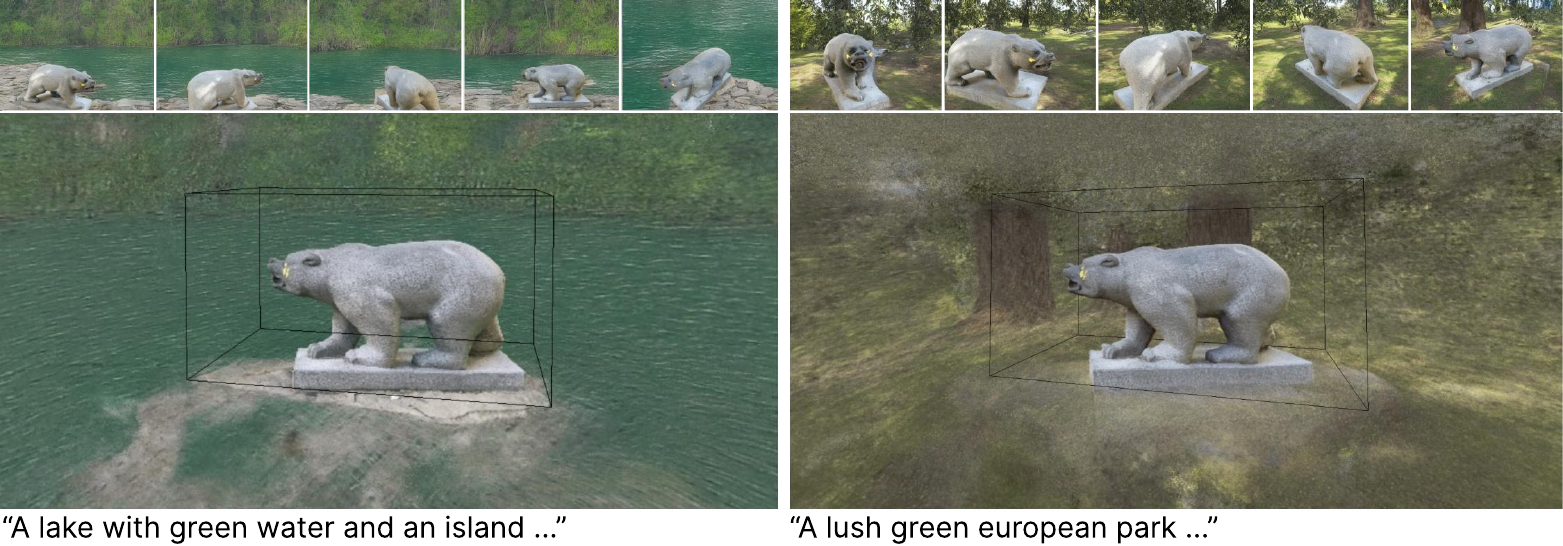}
    \caption{\textbf{Limitations} -- Trying to modify the background by inverting the object mask yields strong inconsistencies. The reference sheet images (top) show non-overlapping views of the background.}
    \label{fig:experiments:limitations}
\end{figure}

    Even though we can achieve better quality than Instruct-NeRF2NeRF with masking, we are forced to downscale the images to fit into the reference sheet passed to the image diffusion model, thereby losing some quality for the generated edit.
    Further, optimal results are obtained when the object occupies the image's center and is positioned close to the camera.
    As increasing the distance reduces the resolution of the object in the latent space the quality of the generated edit diminishes. This behavior holds true for all image diffusion-based 3D generation methods.
    Additionally, an off-center object complicates its incorporation into a reference sheet that generates consistent views, thereby making SIGNeRF unsuitable for extended scene modifications (Fig.~\ref{fig:experiments:limitations}).
\end{subsection}

\section{Conclusion}
\label{sec:conclusion}

With SIGNeRF we present a modular pipeline for scene-integrated editing of NeRF scenes.
An efficient and easily controllable two-step procedure first generates a tiled reference sheet followed by updating the image set to generate consistent edited views suitable for modifying an existing NeRF representation.
The reference sheet additionally shows a preview of the edited scene before generating all images, which is not possible with existing editing methods.
SIGNeRF often achieves consistent 3D generation in a single processing run.
In comparison to previous methods, SIGNeRF is faster and delivers similar or superior editing results without necessitating iterative refinement.

The introduced selection strategies enable generative edits or object insertion within an existing NeRF, even for scenes with complex geometry and appearance.

\ifreview
\else
    \section*{Acknowledgements}
    \vspace{-2mm}
    Funded by EXC number 2064/1 – Project number 390727645. This work was supported by the German Research Foundation (DFG): SFB 1233, Robust Vision: Inference Principles and Neural Mechanisms, TP 02, project number: 276693517.
    The authors thank the International Max Planck Research School for Intelligent Systems (IMPRS-IS) for supporting Jan-Niklas Dihlmann.
\fi

{\small
    \bibliographystyle{ieeenat_fullname}
    \bibliography{11_references}
}

\ifarxiv \clearpage \appendix \renewcommand{\thesection}{\Alph{section}}  %
\renewcommand{\thefigure}{\Alph{section}\arabic{figure}}  %
\renewcommand{\thetable}{\Alph{section}\arabic{table}}    %

This appendix offers supplementary information on SIGNeRF. It delves into more intricate aspects of our implementation (Sec.~\ref{sec:implementation_details}), presents insights into the emergence of multiview consistency (Sec.~\ref{sec:quality_of_multiview_consistency}), outlines the specialized tools we have engineered (Sec.~\ref{sec:viewer}), and elaborates on the generated datasets (Sec.~\ref{sec:dataset_generation}).

\begin{paragraph}{Additional Material}
    We wish to emphasize the inclusion of two videos alongside this paper. Given the three-dimensional nature of our results, these videos serve as the most effective medium for their evaluation. We strongly encourage viewing:

    \begin{itemize}
        \item    \href{https://www.youtube.com/watch?v=TfblNlXNEDc}{\textcolor{blue}{Explanation Video}} | A comprehensive guide through our methods and pipeline.
        \item   \href{https://www.youtube.com/watch?v=BnZzxHEAr0E}{\textcolor{blue}{Results Video}} | A presentation of our results, including comparative analyses.
    \end{itemize}

    \ifreview For a detailed examination at the highest quality, an additional directory named \textcolor{blue}{scene\_videos} has been provided, featuring individual videos of each scene. \fi
    \ifarxiv For a detailed examination and additional information please visit our \href{https://signerf.jdihlmann.com/}{\textcolor{blue}{Project Page}}.\fi
\end{paragraph}

\begin{section}{Implementation Details}
 \label{sec:implementation_details}

 \begin{paragraph}{Image Diffusion}
     In our approach, we integrate an inpainting version of ControlNet~\cite{ControlNet} with Stable Diffusion XL~\cite{SDXL} to synthesize images based on the mask and depth map inputs.
     We have customized the implementation of the publicly available SD WebUI API~\cite{SDWebUI}, which builds upon the Diffusers library~\cite{Diffusers}.
     Notably, SIGNeRF is a general approach that also works with previous versions of Stable Diffusion.
     For example, the `field' scene with the generated cows (Fig.~\ref{fig:experiments:qualitative_results}), which was generated with Stable Diffusion 1.5~\cite{StableDiffusion} and ControlNetInpaint~\cite{ControlNetInpainting}.
 \end{paragraph}

 \begin{paragraph}{Training}
     SIGNeRF necessitates a pre-trained NeRF scene, which we acquire by employing the nerfacto model from Nerfstudio~\cite{Nerfstudio}, undergoing $30,000$ iterations of training for each scene.
     Subsequently, the SIGNeRF pipeline for scene-integrated generation is utilized to create an edited dataset for the NeRF scene~(Sec.~\ref{sec:methods:signerf:scene_integrated_generation}).
     It was discovered that for optimal efficacy, a reference sheet comprising $5$ images strikes an effective balance between preserving image quality and providing adequate scene context.
     Following the creation of the edited dataset, we opt to either fine-tune the existing NeRF scene with the updated dataset or initiate training of a new NeRF scene, depending on the choice of selection method.
     We observe better results for object generation when training the NeRF scene from scratch.
     Conversely, for the task involving generative editing, fine-tuning the pre-trained NeRF scene has shown to be more effective.
     In such cases, optimizers are re-initialized, and the LPIPS~\cite{LPIPS} loss is applied to enhance scene consistency \cite{Instruct-NeRF2NeRF}.
 \end{paragraph}

 \begin{paragraph}{Generation Information}
     \begin{table}
    \centering
    \renewcommand\tabularxcolumn[1]{m{#1}}
    \begin{tabularx}{\linewidth}{
            llll
        }
        \toprule
        {\thead{\textbf{Generation}}} & {\thead{Denoising S.}} & {\thead{ControlNet S.}} & {\thead{Guidance S.}} \\
        \midrule
        Person - Sport                & 0.9                    & 0.4                     & 7.0                   \\
        Person -  Pirate              & [0.95, 0.5]            & [0.4, 0.8]              & 7.5                   \\
        Person -  Batman              & [0.95, 0.5]            & [0.4, 1.0]              & 6.0                   \\
        \midrule
        Plushy - Ironman              & 0.9                    & 1.0                     & 7.0                   \\
        Plushy - Tiger                & 0.6                    & 1.0                     & 7.0                   \\
        Plushy - Gold                 & 0.95                   & 1.0                     & 7.0                   \\
        \midrule
        Bear - Grizzly                & 0.9                    & 0.95                    & 7.0                   \\
        Bear - Polar                  & 0.9                    & 0.95                    & 7.0                   \\
        Bear - Panda                  & 0.9                    & 0.95                    & 7.0                   \\
        Bear - Rabbit                 & 0.95                   & 1.0                     & 7.0                   \\
        \midrule
        Field - Cow                   & (SD 1.5)               & 0.7                     & 7.5                   \\
        \midrule
        Urban - House                 & 1.0                     & 0.8                      & 7.0                    \\
        \bottomrule
    \end{tabularx}
    \caption{\textbf{Scene List} -- Parameters used for each scene for the image diffusion model, with denoising strength, ControlNet scale, and guidance scale.
        Rows with multiple values indicate the need for a second iteration, as discussed in Sec.~\ref{sec:methods:signerf:scene_integrated_generation}.
        For the `field' scene, we used Stable Diffusion 1.5~\cite{StableDiffusion}, which does not have a denoising strength parameter.
    }
    \label{tbl:scene_parameters}
\end{table}

     The generational edit of the NeRF scene can be controlled by several parameters provided by the image diffusion model, such as the denoising strength, the ControlNet condition and the guidance scale.
     We provide the parameters used for each scene in Tbl.~\ref{tbl:scene_parameters}.
 \end{paragraph}

 \begin{paragraph}{Metrics}
     For the quantitative evaluation of our results (Sec.~\ref{sec:experiments:quantitative}), we use two metrics, the first one is the CLIP text-to-image similarity score~\cite{StyleGANNADA} and the second one is a new background preservation metric.
     This background preservation metric measures the background difference between a render~$\boldsymbol{I_{\text{org}}}$ from the original NeRF and a render~$\boldsymbol{I_{\text{edit}}}$ from the edited NeRF.
     We use the corresponding mask~$\boldsymbol{M}$ provided by the picked selection method~(Sec.~\ref{sec:method:selection}) and compute the background preservation as:
     \begin{equation}
         \text{PSNR}_\text{bg} = \text{PSNR}(\boldsymbol{I_{\text{org}}}  \cdot (\neg\boldsymbol{M}), \boldsymbol{I_{\text{edit}}}  \cdot (\neg\boldsymbol{M}))
     \end{equation}
     \begin{equation}
         \text{SSIM}_\text{bg} = \text{SSIM}(\boldsymbol{I_{\text{org}}}  \cdot (\neg\boldsymbol{M}), \boldsymbol{I_{\text{edit}}}  \cdot (\neg\boldsymbol{M}))
     \end{equation}
 \end{paragraph}
\end{section}

\begin{section}{Multiview Consistency}
 \label{sec:quality_of_multiview_consistency}
 \captionsetup[table]{font=small,skip=0pt}
\begin{figure}[t!] %
    \centering
    \includegraphics[width=\linewidth]{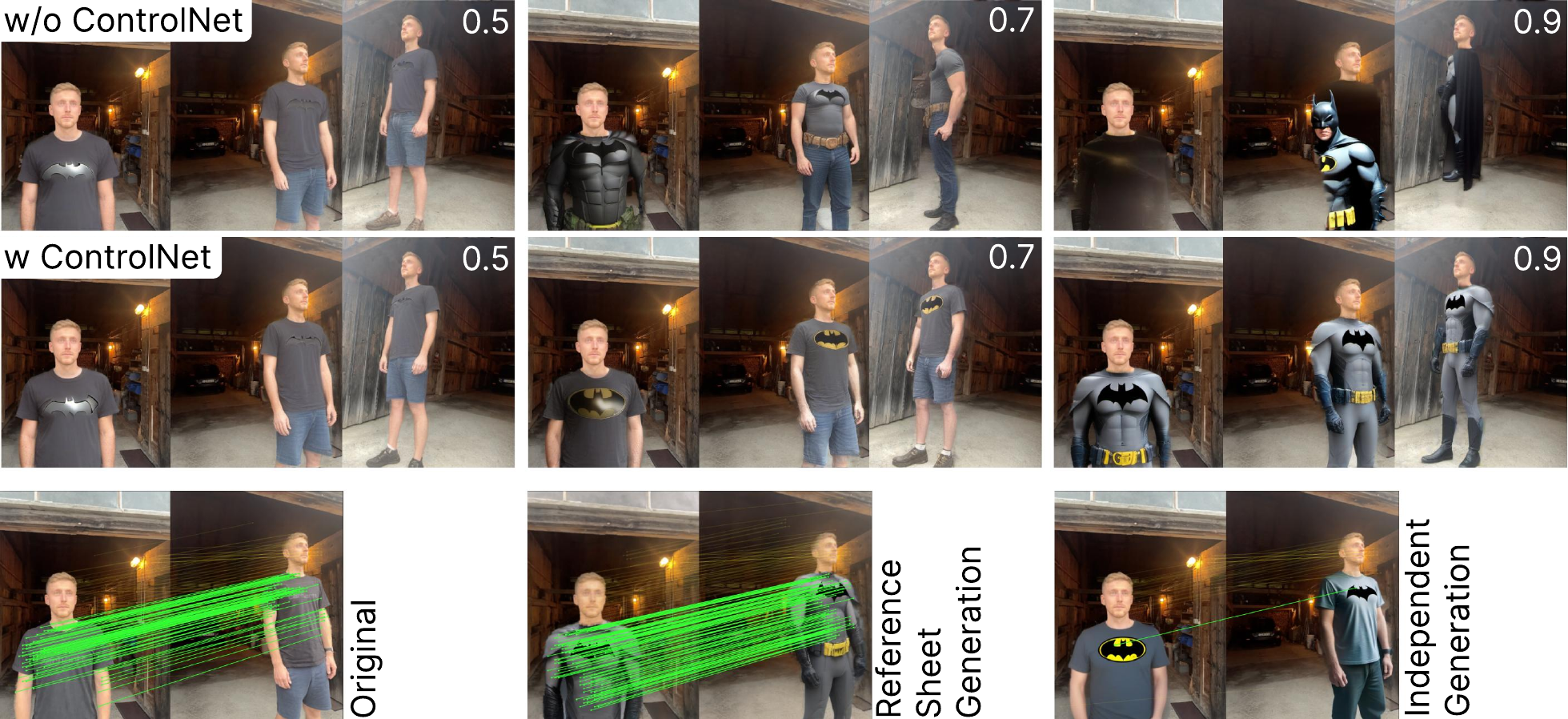}
    \caption{ \small \textbf{Inpainting Ablation \& LoFTR Features} -- grid inpainting generation capabilities with \& w/o ControlNet under certain denoising levels (label top right). LoFTR keypoint correspondences reflect the consistency between the sub images (last row). Zoom in for details.}
    \label{fig:rebuttal:gridInpainting}
    \vspace*{-1.5em}
\end{figure}

 We want to clarify that ControlNet~\cite{ControlNet} was not fine-tuned in any way.
 We found the multi-view consistency property empirically by exploration.
 Initially, combining two images, and later discovering how to reach greater consistency using a grid.
 The reason why diffusion models show this multi-view consistency isn't entirely clear, but we have some hypotheses.
 The grid-based inpainting can achieve a degree of multi-view consistency without ControlNet, although the results become less consistent at higher denoising levels (Fig. \ref{fig:rebuttal:gridInpainting}), therefore only allowing subtle edits.
 This suggests that StableDiffusion~\cite{StableDiffusion} inherently supports some level of multi-view consistency and 3D understanding as also shown by~\cite{chen2023beyond, zhan2023does}.
 We assume that attention plays a crucial role, further enforced by the inpainting mechanism focusing on the relevant parts and ignoring scene elements by masking.
 The depth condition in ControlNet further acts as a reference for the model to infer orientation/shape/scale, even for high denoising levels as the condition is not noised.
 Fine-tuning to achieve even better results is interesting future work, but we think that one of the most attractive properties of our approach is the possibility of achieving high-quality generation with off-the-shelf image diffusion models.

 Further there is no direct measurement of the quality of the multiview consistency within the reference sheet.
 This is also nontrivial to measure, else we could optimize for it in an self supervised manner.
 We experimented with LoFTR~\cite{sun2021loftr} correspondences between original, reference-sheet generated and independent generated images.
 Averaged results show that independent generation has $70\%$ less correspondences than the original images, while the reference sheet generated images achieve $15\%$ less correspondences.
 Nevertheless, we believe the supplementary videos we provided substantiate our claims of achieving multiview consistency.
 The feasibility of conducting 3D reconstructions from the generated images serves as a testament to this consistency.
 It is important to acknowledge that while the images may not exhibit perfect consistency, the underlying  NeRF counteracts that by aligning 3D consistent regions and discarding outliers.
\end{section}

\begin{section}{Viewer}
 \label{sec:viewer}
 Our development includes a NeRF viewer designed to examine the trained NeRF scenes and facilitate the placement of proxy objects or the selection of scene elements for modification. The viewer's backend is based on Nerfstudio \cite{Nerfstudio}, while the frontend is entirely novel, incorporating a suite of advanced features. A depiction of the viewer interface is presented in Fig.~\ref{fig:appendix:viewer}. Constructed as a React \cite{React} application, it leverages React Three Fiber \cite{R3F}, a React renderer for Three.js \cite{Three}, to facilitate 3D scene manipulation. The Nerfstudio backend streams the NeRF data to the frontend, where it is displayed centrally in the viewport. We have augmented the backend to transmit not just the scene render but also the corresponding depth map. This data enables us to utilize a shader that merges the NeRF render with Three.js's native rendering, allowing for the visualization of occluded elements within the NeRF scene in real-time.
 \begin{figure}[t!]
    \centering
    \includegraphics[width=\linewidth]{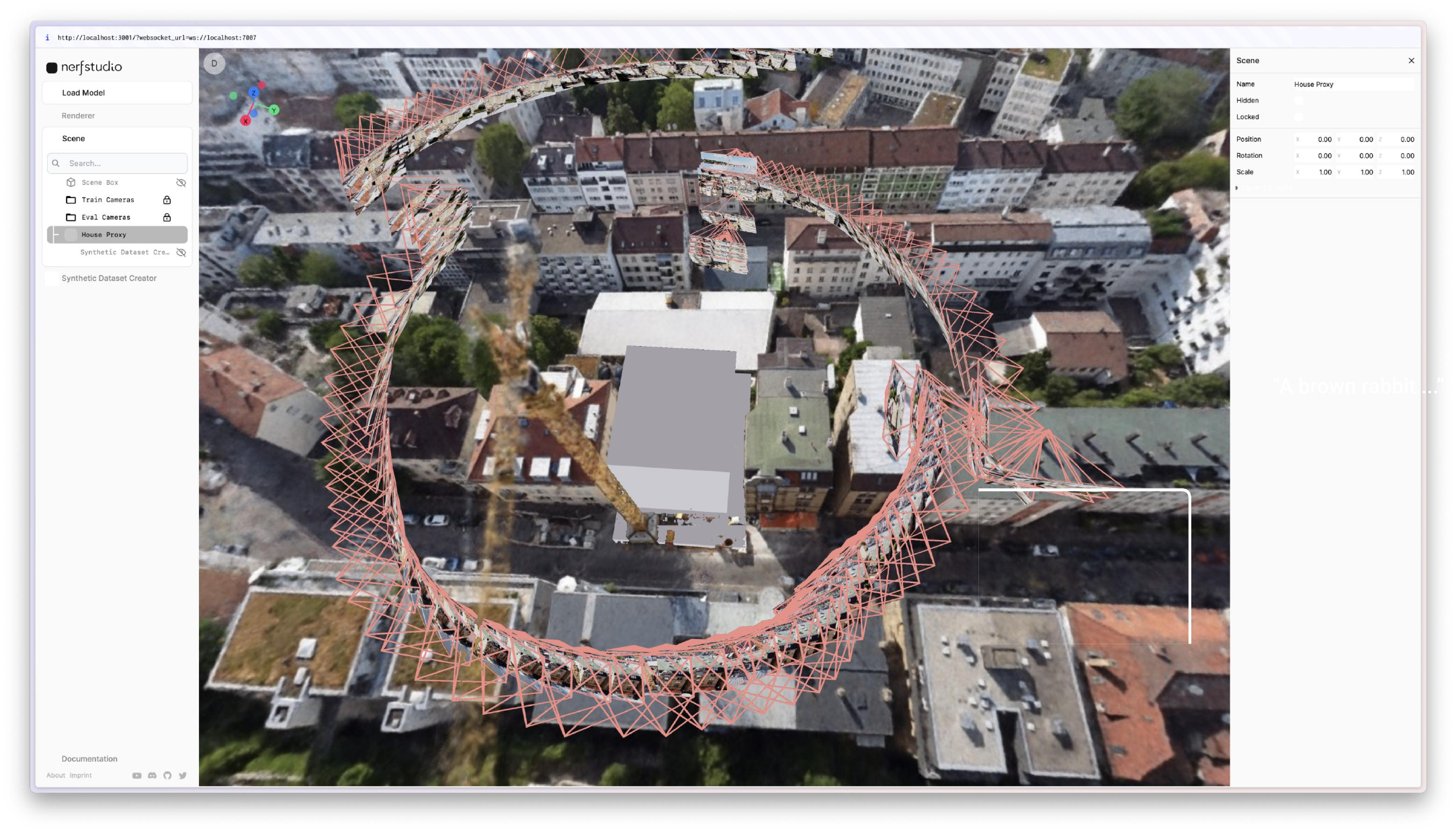}
    \caption{\textbf{Viewer} -- Outlining the placement of a proxy object within an existing NeRF scene. The viewer is divided into a 3D viewport at the center, a scene and selection column on the left and a selection specific control column on the right.
    }
    \label{fig:appendix:viewer}
\end{figure}

 Enhancements to the viewer include a scene hierarchy and intuitive manipulation tools, which simplify the process of placing proxy objects or delineating bounding boxes to select regions of interest. Additionally, a bridge to SIGNeRF was written, such that we can generate the reference sheet and dataset with the information provided by the viewer. Consequently, the viewer serves as a user-friendly platform for performing scene-integrated generation on NeRF scenes. We plan to release this viewer as open-source software, providing the community with a powerful tool for NeRF scene editing.
\end{section}

\begin{section}{Dataset Generation}
 \label{sec:dataset_generation}
 \begin{figure*}[t!]
    \centering
    \includegraphics[width=\textwidth]{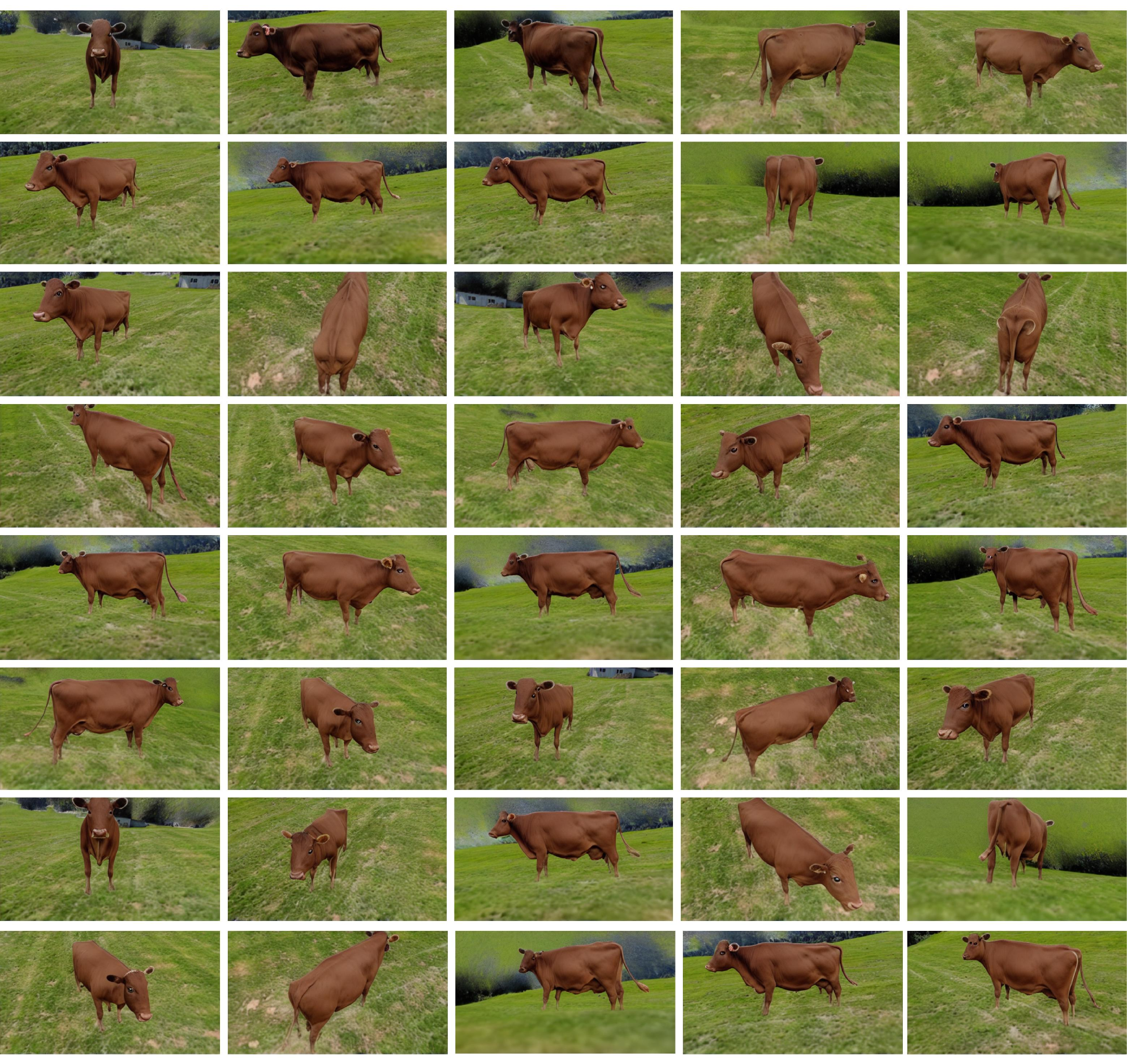}
    \caption{\textbf{Dataset Generation} -- Illustrating the consistency within edited views of an original NeRF dataset.
        The first row shows images from the reference sheet, the rest are the views generated with the reference sheet.
    }
    \label{fig:appendix:dataset_generation}
\end{figure*}

 SIGNeRF generates an updated NeRF image dataset with the provided reference sheet to edit a NeRF scene.
 Our generative pipeline~(Fig.~\ref{fig:methods:pipeline}) demonstrates this with a selection of modified dataset images, but a completely revised dataset is not showcased.
 In Fig.~\ref{fig:appendix:dataset_generation}, we exhibit a fully edited dataset of the `field' scene (Fig.~\ref{fig:experiments:qualitative_results}).
 Consistency is maintained across the generated images; however, some outliers might occur.
 These inconsistencies are minor over the dataset and are generally resolved by the NeRF optimization process, which learns to ignore such outliers.
 For complex scenes, the use of an LPIPS~\cite{LPIPS} loss is beneficial to further improve the consistency during the optimization.
\end{section}

 \fi

\end{document}